\documentclass{article}

\usepackage{arxiv}

\usepackage[utf8]{inputenc} 
\usepackage[T1]{fontenc}    
\usepackage{hyperref}       
\usepackage{url}            
\usepackage{booktabs}       
\usepackage{amsfonts}       
\usepackage{nicefrac}       
\usepackage{microtype}      
\usepackage{lipsum}		
\usepackage{graphicx}
\usepackage{natbib}
\usepackage{doi}
\usepackage{subcaption}
\usepackage{placeins}
\usepackage{caption}
\title{MiniGPT: Rebuilding GPT from First Principles}

\author{
\href{https://orcid.org/0009-0001-8585-2354}
{\includegraphics[scale=0.06]{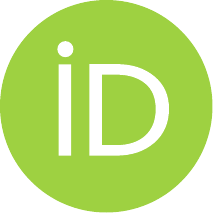}\hspace{1mm}Jibin~Joseph} \\
Department of Computer Science \\
The University of Texas at Austin \\
Austin, TX 78712 \\
\texttt{jibinjoseph@utexas.edu}
}

\renewcommand{\shorttitle}{MiniGPT}

\hypersetup{
pdftitle={MiniGPT: Rebuilding GPT from First Principles},
pdfsubject={cs.LG, cs.CL},
pdfauthor={Jibin Joseph},
pdfkeywords={MiniGPT, GPT, Transformer, autoregressive language model, causal self-attention, PyTorch, Tiny Shakespeare, character-level language modeling},
}

\begin{document}
\maketitle

\begin{abstract}
This paper presents MiniGPT, a compact from-scratch implementation of GPT-style autoregressive language modeling in PyTorch. The aim is to rebuild the core GPT pipeline from first principles after studying the design of nanoGPT by Andrej Karpathy, while keeping the model and training code independently written in a single notebook. MiniGPT implements token and positional embeddings, causal multi-head self-attention, pre-LayerNorm Transformer blocks, residual connections, feed-forward MLP layers, next-token cross-entropy training (teacher forcing), validation tracking, checkpoint selection, and autoregressive text generation. This paper evaluates the implementation on Tiny Shakespeare dataset using character-level tokenization. A baseline 0.83M-parameter model reaches a validation loss of 1.7236 after 3000 training iterations. A stronger 10.77M-parameter configuration, using a larger context length and improved training settings, reaches a best validation loss of 1.4780 and generates text with recognizable Shakespeare-style dialogue structure. MiniGPT does not introduce a new language-model architecture. Instead, it documents a clear and reproducible implementation path from raw text to trained character-level generation, including design choices, training behavior, generation quality, and practical limitations.
\end{abstract}


\section{Introduction}

Generative language models have become one of the most visible outcomes of modern deep learning. Modern deep learning is often described as representation learning with multilayer neural networks \citep{lecun2015deep}, while broader historical accounts show that deep learning also has a longer lineage in neural networks, credit assignment, supervised learning, unsupervised learning, and reinforcement learning \citep{schmidhuber2015deep}. In language modeling, this idea becomes especially important because a model can learn from raw text by predicting the next token in a sequence. The Transformer architecture made this approach more practical by using attention instead of recurrence or convolution as the main sequence-modeling mechanism \citep{vaswani2017attention}. GPT-style models build on this idea by using a decoder-only Transformer trained with an autoregressive next-token prediction objective \citep{radford2018improving,radford2019language}. Later large language models showed that the same basic training idea can scale to much larger models and broader language tasks \citep{brown2020language}.

However, the core GPT pipeline can still be difficult to understand from high-level explanations alone. Many descriptions show the model as a diagram, while many real implementations include engineering details that are useful for large-scale training but harder to follow at first. Compact implementations such as minGPT and nanoGPT helped make GPT-style modeling easier to inspect in code \citep{karpathy2020mingpt,karpathy2022nanogpt}. nanoGPT is especially useful as a practical reference because it organizes the model definition, training loop, evaluation, checkpointing, and sampling workflow in a small codebase \citep{karpathy2022nanogpt}. Still, rebuilding the same core ideas in a single notebook can make the full path from raw text to generated samples more direct and easier to reproduce.

This paper presents MiniGPT, a compact from-scratch implementation of GPT-style autoregressive language modeling in PyTorch. MiniGPT follows the high-level decoder-only Transformer design, but the model and training code are written independently. The implementation includes token embeddings, positional embeddings, masked multi-head self-attention, feed-forward Transformer blocks, residual connections, layer normalization, next-token cross-entropy training, validation-loss tracking, checkpoint selection, and autoregressive text generation. The goal is not to introduce a new language-model architecture. Instead, the goal is to document a clear and reproducible implementation path for a small GPT-style model.

MiniGPT is evaluated on Tiny Shakespeare dataset using character-level tokenization. This setting is intentionally small, but it is useful because the dataset contains repeated structure such as speaker names, punctuation, line breaks, and dialogue-like text. The implementation first trains a baseline model to verify the complete pipeline, and then trains a stronger configuration to study the effect of increased model capacity, longer context length, validation-based checkpoint selection, and autoregressive generation quality.

The main contributions of this paper are:

\begin{itemize}
    \item A from-scratch PyTorch implementation of a small decoder-only GPT-style language model.
    \item A complete character-level language-modeling pipeline from raw text to train and validation batches.
    \item Baseline and stronger Tiny Shakespeare dataset experiments with loss tracking and checkpoint selection.
    \item Qualitative generation analysis using autoregressive sampling.
    \item A practical discussion of model size, context length, tokenization, overfitting, and sampling behavior.
\end{itemize}

MiniGPT is a technical implementation and reproducibility paper. It does not claim architectural novelty or state-of-the-art language modeling performance. Its value is in making the main GPT-style training pipeline explicit, compact, and reproducible.

\section{Background and Related Work}

\subsection{Transformer Language Models}

Language modeling is the task of learning patterns in text so that a model can predict upcoming tokens from previous context. Earlier neural language models showed that distributed representations can reduce the sparsity problem in word-sequence modeling \citep{bengio2003neural}. The Transformer later changed sequence modeling by replacing recurrence and convolution with attention-based layers \citep{vaswani2017attention}. This made it possible to build models where each token representation is computed from other tokens in the sequence through self-attention.

GPT-style language models use the Transformer idea in a decoder-only setting. In this setup, the model reads only the previous context and learns to predict the next token. This autoregressive training objective was used in the original GPT work for generative pre-training \citep{radford2018improving}. GPT-2 showed that the same basic objective can support broader language behavior when scaled to larger models and larger text datasets \citep{radford2019language}. GPT-3 further showed that large autoregressive language models can perform many tasks through prompting and in-context examples, without updating model parameters for each task \citep{brown2020language}.

MiniGPT follows the same general next-token prediction idea, but it keeps the setting intentionally small. Instead of focusing on scale, it focuses on making the main parts of a GPT-style model visible.

\subsection{Tokenization and Small Language-Model Settings}
\label{sec:tokenization}
Tokenization is an important design choice in language modeling. Many larger language models use subword or byte-level tokenization because it gives a practical balance between word-level meaning and character-level flexibility. Subword methods such as byte pair encoding are useful because they can represent rare or unseen words as smaller units \citep{sennrich2016neural}. GPT-2 also used a byte-level byte pair encoding approach as part of its language modeling pipeline \citep{radford2019language}.

MiniGPT uses character-level tokenization instead. Character-level language modeling with self-attention has also been studied in prior work, showing that deep attention-based models can learn useful structure directly from character sequences \citep{alrfou2018character}. This choice is simpler than subword tokenization and keeps the full preprocessing pipeline easy to inspect. Each unique character becomes one token, so the vocabulary is small and no external tokenizer is required. The trade-off is that the model must learn words, punctuation, and longer patterns from individual characters, Which makes long-range coherence harder than with subword tokenization \citep{sennrich2016neural}.

\subsection{nanoGPT as a Design Reference}

Several implementation-oriented resources have made GPT-style models easier to study in code. \citet{karpathy2020mingpt} provides a minimal PyTorch reimplementation of GPT focused on small, readable model code. \citet{karpathy2022nanogpt} extends this direction with a compact but more practical GPT training repository. nanoGPT is especially relevant because it includes a character-level Shakespeare example, model definition, training loop, validation evaluation, checkpointing, and sampling workflow \citep{karpathy2022nanogpt}.

MiniGPT uses nanoGPT as a design reference, not as copied or imported code. The aim is to rebuild the same core ideas in a single notebook: a decoder-only Transformer, causal masking, next-token training, and autoregressive generation. This keeps the implementation smaller and more direct than a full training repository, while still preserving the main GPT-style workflow.

There are also larger Transformer software ecosystems, such as the Hugging Face Transformers library, which provide many pretrained models and unified APIs for training, fine-tuning, and inference \citep{wolf2020transformers}. MiniGPT is different from these libraries. It is not a pretrained-model platform and it is not meant to cover many architectures. It is a compact implementation report that shows the smallest complete path from raw text to trained character-level generation.

\section{MiniGPT Architecture}

MiniGPT is implemented as a small decoder-only Transformer language model. The model follows the standard GPT-style structure: token IDs are converted into token embeddings, positional information is added, the sequence is processed by a stack of Transformer blocks, and a final linear language-model head produces logits over the vocabulary. This follows the general Transformer design introduced by \citet{vaswani2017attention} and the decoder-only autoregressive language-modeling direction used in GPT-style models \citep{radford2018improving,radford2019language}. The implementation is intentionally compact and written from scratch in PyTorch, while using nanoGPT only as a high-level design reference \citep{karpathy2022nanogpt}.

\subsection{Model Overview}

The high-level architecture of MiniGPT is:

\begin{center}
\texttt{Token IDs} $\rightarrow$ \texttt{Token Embedding + Positional Embedding} $\rightarrow$ \texttt{Transformer Blocks} $\times L$ $\rightarrow$ \texttt{Final LayerNorm} $\rightarrow$ \texttt{Linear LM Head}
\end{center}

The model receives a batch of token IDs with shape $(B, T)$, where $B$ is the batch size and $T$ is the sequence length. Each token ID is mapped to a learned vector using a token embedding table. A learned positional embedding is also added so that the model can distinguish the order of tokens in the sequence. This is necessary because self-attention alone does not directly encode token position \citep{vaswani2017attention}.

The main architecture settings are stored in a configuration object. This includes the context length, vocabulary size, number of Transformer layers, number of attention heads, embedding dimension, and dropout probability. In the baseline configuration, MiniGPT uses a context length of 128, 4 Transformer layers, 4 attention heads, and an embedding dimension of 128. This gives a small model with 826,433 trainable parameters. In the stronger configuration, MiniGPT uses a context length of 256, 6 Transformer layers, 6 attention heads, and an embedding dimension of 384, giving 10.77M trainable parameters (Table~\ref{tab:training_pipeline}, Table~\ref{tab:experiment_results})).

\subsection{Causal Self-Attention}

The core operation in MiniGPT is causal multi-head self-attention. For each input sequence, the model projects the hidden states into query, key, and value vectors. These vectors are split across multiple attention heads, allowing the model to learn different token-to-token relationships in parallel. The attention scores are computed using scaled dot-product attention, following the Transformer formulation \citep{vaswani2017attention}.

For one attention head, this can be written as
\[
\mathrm{Attention}(Q,K,V)=\mathrm{softmax}\left(\frac{QK^\top}{\sqrt{d_{\mathrm{head}}}} + M\right)V,
\]
where \(M\) is the causal mask. In MiniGPT, \(M_{ij}=0\) when \(i \geq j\), and \(M_{ij}=-\infty\) when \(i<j\), so future positions are blocked before the softmax.

Because MiniGPT is an autoregressive language model, each token is only allowed to attend to itself and earlier tokens. It should not attend to future tokens, because the future token is the target that the model is trying to predict. MiniGPT enforces this using a lower-triangular causal mask. The mask blocks all future positions before the softmax operation. This makes the model suitable for next-token prediction, because the prediction at each position depends only on the available left context.

In implementation terms, the attention module applies the following steps: project inputs into queries, keys, and values; reshape them into multiple heads; compute scaled attention scores; apply the causal mask; apply softmax; combine the values; merge the heads; and apply a final output projection. This keeps the implementation close to the mathematical idea of self-attention and makes the masking behavior easy to inspect.

\subsection{Transformer Block}

Each MiniGPT Transformer block contains two main submodules: causal self-attention and a feed-forward MLP. The block uses residual connections around both submodules. Residual connections are useful because they make deeper neural networks easier to optimize by allowing layers to learn changes relative to their inputs \citep{he2016deep}. In MiniGPT, the block has the following form:

\begin{center}
\texttt{x = x + Attention(LayerNorm(x))} \\
\texttt{x = x + MLP(LayerNorm(x))}
\end{center}

This is a pre-LayerNorm design, where layer normalization is applied before the attention and MLP submodules. Layer normalization stabilizes hidden-state statistics during training \citep{ba2016layer}. The pre-LayerNorm placement is also commonly used in practical Transformer implementations because it can make optimization more stable, especially as models become deeper \citep{xiong2020layer}.

The feed-forward MLP expands the embedding dimension by a factor of four and then projects it back to the original embedding dimension. MiniGPT uses the structure:

\begin{center}
\texttt{Linear($d$, 4$d$) $\rightarrow$ GELU $\rightarrow$ Linear(4$d$, $d$) $\rightarrow$ Dropout}
\end{center}

where $d$ is the embedding dimension. GELU is used as the activation function because it is commonly used in Transformer language models and provides a smooth nonlinear transformation \citep{hendrycks2016gelu}. Dropout is included as a regularization method, especially because Tiny Shakespeare is a small dataset and overfitting can happen easily \citep{srivastava2014dropout}.

\subsection{Language-Model Head and Loss}

After the final Transformer block, MiniGPT applies a final LayerNorm and then a linear language-model head. The language-model head maps each hidden state to a vector of logits over the vocabulary. For a character-level Tiny Shakespeare model, the vocabulary size is 65, so each position produces 65 logits.

When target tokens are provided, MiniGPT computes the cross-entropy loss between the predicted logits and the next-token targets. The target sequence is the input sequence shifted by one position. This means that at every position, the model learns to predict the next character from the previous context. This is the same basic autoregressive objective used in GPT-style language modeling \citep{radford2018improving,radford2019language}.

In the stronger MiniGPT configuration, the token embedding matrix and language-model head weights are tied when their shapes match. Weight tying uses the same matrix for input token embeddings and output prediction weights. This can reduce the number of trainable parameters and has been shown to improve language-model performance in some settings \citep{press2017using}. Here, \(V=65\) and \(d_{\mathrm{embd}}=384\), so tying the token embedding and language-model head weights saves \(65 \times 384 = 24{,}960\) parameters. In MiniGPT, this is used as a simple practical improvement while keeping the architecture easy to understand.

\section{Training Pipeline}

MiniGPT uses a simple training pipeline for character-level autoregressive language modeling. The goal of the pipeline is to convert raw text into token sequences, form next-token prediction examples, train the model with cross-entropy loss, and track both training and validation loss over time. This follows the standard autoregressive language-modeling setup used in GPT-style models, where the model learns to predict the next token from the previous context \citep{radford2018improving,radford2019language}.

\subsection{Dataset and Tokenization}

The experiments use the Tiny Shakespeare, a small text dataset commonly used for character-level language modeling examples \citep{karpathy2015charRNN}. The dataset is small enough to train quickly in a notebook environment, but it still contains useful structure such as speaker names, line breaks, punctuation, and dialogue-like text.

MiniGPT uses character-level tokenization as discussed in Section~\ref{sec:tokenization}. In this setup, each unique character in the dataset is treated as one token. The tokenizer builds two dictionaries: one mapping characters to integer token IDs, and another mapping token IDs back to characters. This gives a vocabulary size of 65 characters.

After tokenization, the full text is converted into one long tensor of integer token IDs. The data is then split into training and validation sets using a 90/10 split. The training split is used for parameter updates, while the validation split is used only to estimate how well the model is doing on unseen text.

\subsection{Input-Target Construction}

For next-token prediction, each training example is built from a fixed-length block of tokens. Given a starting index $i$ and a context length called \texttt{block\_size}, the input sequence is:

\begin{center}
\texttt{x = tokens[i : i + block\_size]}
\end{center}

and the target sequence is the same window shifted one token to the right:

\begin{center}
\texttt{y = tokens[i + 1 : i + block\_size + 1]}.
\end{center}

This means that at every position, the model receives the current and previous characters and learns to predict the next character. For example, if the input is \texttt{hell}, the target is \texttt{ello}. This shifted input-target construction is the core of autoregressive language modeling.

MiniGPT samples random starting positions from either the training split or the validation split. For each minibatch, it stacks multiple independent token blocks into tensors of shape $(B, T)$, where $B$ is the batch size and $T$ is the context length. These tensors are moved to the selected device, either CPU or GPU, before the forward pass.

\subsection{Optimization and Validation}
\label{sec:optimization}
MiniGPT is trained using cross-entropy loss between the predicted vocabulary logits and the shifted target tokens. The logits are reshaped from $(B, T, V)$ to $(B \times T, V)$, where $V$ is the vocabulary size, and the targets are reshaped from $(B, T)$ to $(B \times T)$. This lets the model compute one next-token prediction loss for every position in every sequence.

The baseline model uses AdamW with a fixed learning rate. AdamW is used because it applies decoupled weight decay, which is a practical improvement over standard Adam-style weight decay handling \citep{loshchilov2019decoupled}. Adam itself is based on adaptive estimates of first and second moments of gradients, which makes it a widely used optimizer for neural network training \citep{kingma2015adam}.

The baseline training run uses a small configuration: batch size 32, context length 128, learning rate $3 \times 10^{-4}$, and 3000 training iterations (Table~\ref{tab:training_pipeline}). Periodically, MiniGPT estimates both training loss and validation loss by averaging over several minibatches. This makes the loss curve less noisy than reporting a single batch loss. The results are presented in the Section~\ref{sec:results} (Table~\ref{tab:experiment_results}). 

A stronger MiniGPT configuration is also trained to improve generation quality. This run uses a larger model, context length 256, batch size 64, dropout 0.2, AdamW with parameter groups, warmup followed by cosine learning-rate decay, gradient clipping, mixed precision training, and checkpoint selection based on validation loss. The stronger run uses a maximum learning rate of \(10^{-3}\), minimum learning rate of \(10^{-4}\), 100 warmup steps, cosine decay over 5000 steps, AdamW betas of \((0.9, 0.99)\), weight decay of \(10^{-1}\), and gradient clipping at 1.0 (Table~\ref{tab:training_pipeline}). For AdamW parameter grouping, tensors with dimension two or higher use weight decay, while one-dimensional tensors such as biases and LayerNorm parameters use no weight decay; in the implementation this gives 38 decayed parameter tensors and 63 non-decayed parameter tensors. Cosine-style learning-rate schedules are commonly used to reduce the learning rate smoothly during training \citep{loshchilov2017sgdr}. Gradient clipping is used to limit unusually large gradients and improve training stability \citep{pascanu2013difficulty}. Mixed precision is used on GPU to reduce memory use and improve training speed while keeping training stable with loss scaling \citep{micikevicius2018mixed}.

\begin{table}[h]
\centering
\renewcommand{\arraystretch}{1.20}
\begin{tabular}{lcc}
\hline
\textbf{Setting} & \textbf{Baseline MiniGPT} & \textbf{Stronger MiniGPT} \\
\hline
Dataset & Tiny Shakespeare & Tiny Shakespeare \\
Tokenization & Character-level & Character-level \\
Vocabulary size & 65 & 65 \\
Train/validation split & 90/10 & 90/10 \\
Batch size & 32 & 64 \\
Context length & 128 & 256 \\
Training iterations & 3000 & 5000 \\
Optimizer & AdamW & AdamW \\
Learning-rate schedule & Fixed & Warmup + cosine decay \\
Gradient clipping & No & Yes \\
Mixed precision & No & Yes, on CUDA \\
Checkpoint selection & Final model & Best validation loss \\
Maximum learning rate & $3 \times 10^{-4}$ & $10^{-3}$ \\
Minimum learning rate & -- & $10^{-4}$ \\
Warmup steps & -- & 100 \\
Cosine decay steps & -- & 5000 \\
AdamW betas & Default & $(0.9, 0.99)$ \\
Weight decay & Default AdamW setting & $10^{-1}$ for decayed parameters \\
Parameter grouping & No & 38 decayed / 63 non-decayed tensors \\
\hline
\end{tabular}
\caption{Summary of the MiniGPT training pipeline settings.}
\label{tab:training_pipeline}
\end{table}

\section{Experiments and Results}
\label{sec:results}
This section reports two MiniGPT experiments on the Tiny Shakespeare dataset as discussed in Section~\ref{sec:optimization}. The first experiment uses a small baseline model to verify that the full training pipeline works correctly. The second experiment uses a stronger model configuration to improve validation loss and generation quality. The goal is not to reach state-of-the-art language modeling performance, but to show that the implementation learns meaningful structure from raw text.

\subsection{Baseline Configuration}

The baseline experiment uses a small decoder-only Transformer with 4 layers, 4 attention heads, an embedding dimension of 128, and a context length of 128 characters. This model has 826,433 trainable parameters. It is trained for 3000 iterations using AdamW and character-level next-token prediction. AdamW is used because it is a standard optimizer for Transformer training and applies decoupled weight decay \citep{loshchilov2019decoupled}.

At the start of training, both training and validation loss are close to 4.20. This is expected because the model starts with random weights and the Tiny Shakespeare vocabulary has 65 characters. During training, both losses decrease steadily. By step 3000, the baseline model reaches a training loss of 1.5304 and a validation loss of 1.7236 (Table~\ref{tab:experiment_results}, Figure~\ref{fig:strong_loss_curve}). This confirms that the model learns useful local patterns from the text and that the training loop, validation loop, and loss computation are working correctly.

\subsection{Stronger Configuration}

The stronger experiment increases the model capacity and training quality. It uses 6 Transformer layers, 6 attention heads, an embedding dimension of 384, and a context length of 256 characters. This gives a 10.77M-parameter model. The stronger configuration also uses dropout, AdamW parameter groups, warmup followed by cosine learning-rate decay, gradient clipping, mixed precision training on CUDA, and checkpoint selection based on validation loss.

This configuration is close to the small Shakespeare character-level setup used in nanoGPT, where a 6-layer, 6-head, 384-channel GPT is trained on Tiny Shakespeare dataset \citep{karpathy2022nanogpt}. MiniGPT does not reuse nanoGPT code, but this comparison is useful because it gives a practical reference point for the scale of the experiment.

The stronger model starts with a validation loss of 4.2879. The loss then drops quickly during the early part of training. The best validation loss is 1.4780 at step 1750. The corresponding validation perplexities are approximately 5.60 for the baseline model and 4.38 for the stronger best checkpoint (Table~\ref{tab:experiment_results}). After this point, the training loss continues to decrease, but the validation loss begins to increase (Figure~\ref{fig:strong_loss_curve}). This shows that the model starts to overfit the training text after the best checkpoint. Validation-based checkpoint selection is therefore important, because the final model at step 5000 is not the best model for generalization \citep{prechelt1998early}.

\begin{table}[!htbp]
\centering
\renewcommand{\arraystretch}{1.15}
\begin{tabular}{lcc}
\hline
\textbf{Metric} & \textbf{Baseline MiniGPT} & \textbf{Stronger MiniGPT} \\
\hline
Parameters & 826,433 & 10.77M \\
Layers / heads / embedding dim & 4 / 4 / 128 & 6 / 6 / 384 \\
Context length & 128 & 256 \\
Batch size & 32 & 64 \\
Training iterations & 3000 & 5000 \\
Checkpoint used & Final step & Best validation checkpoint \\
Best/final training loss & 1.5304 & 1.0990 at step 1750 \\
Best/final validation loss & 1.7236 at step 3000 & 1.4780 at step 1750 \\
Validation perplexity, $\exp(\mathrm{loss})$ & 5.60 & 4.38 \\
Final validation loss & 1.7236 & 1.7055 at step 5000 \\
Training time & 50.79 seconds & 4.76 minutes \\
\hline
\end{tabular}
\caption{Comparison of baseline and stronger MiniGPT experiments on Tiny Shakespeare dataset. Perplexity is computed as $\exp(\mathrm{validation\ loss})$ for the reported validation checkpoint.}
\label{tab:experiment_results}
\end{table}

\subsection{Loss Curves and Checkpoint Selection}
\label{sec:losscurves}

\begin{figure}[!htbp]
\centering
\includegraphics[width=0.80\linewidth]{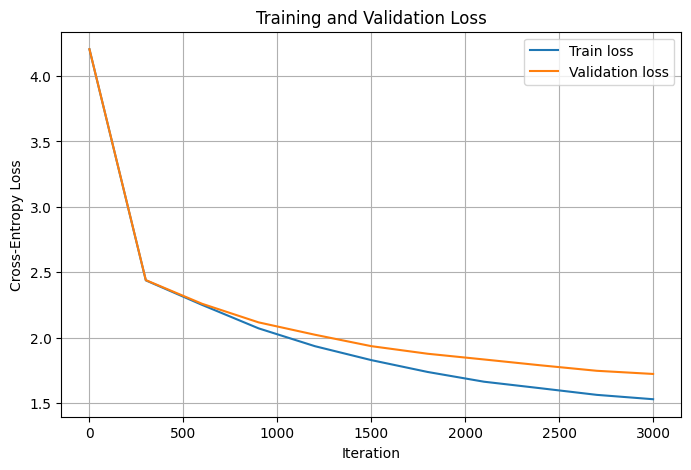}
\caption{Training and validation loss for the baseline MiniGPT model on Tiny Shakespeare dataset. Both losses decrease steadily from about 4.20 at step 0 to 1.5304 training loss and 1.7236 validation loss at step 3000, showing that the baseline model learns useful structure from the dataset without clear overfitting.}
\label{fig:baseline_loss_curve}
\end{figure}

\begin{figure}[!htbp]
\centering
\includegraphics[width=0.80\linewidth]{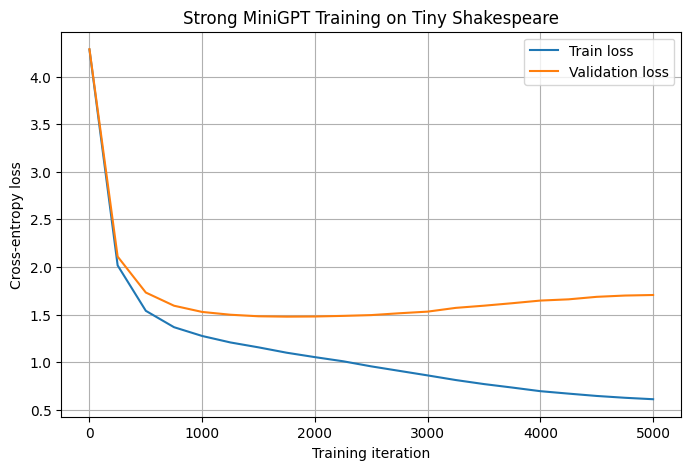}
\caption{Training and validation loss for the stronger MiniGPT model on Tiny Shakespeare dataset. The validation loss reaches its minimum value of 1.4780 at step 1750, while the training loss continues to decrease afterward. By step 5000, the training loss falls to 0.6105, but the validation loss increases to 1.7055, indicating overfitting after the best validation checkpoint (step 1750).}
\label{fig:strong_loss_curve}
\end{figure}

The loss curves show two important behaviors. First, the stronger model learns faster and reaches a lower validation loss than the baseline model. The baseline model reaches 1.7236 validation loss after 3000 steps (Figure~\ref{fig:baseline_loss_curve}), while the stronger model reaches 1.4780 validation loss by step 1750 (Figure~\ref{fig:strong_loss_curve}). This improvement is mainly due to the larger model capacity, longer context length, stronger regularization, learning-rate scheduling, and better checkpoint selection.

Second, the stronger model clearly shows overfitting after step 1750. At step 1750, the training loss is 1.0990 and the validation loss is 1.4780. By step 5000, the training loss has decreased further to 0.6105, but the validation loss has increased to 1.7055 (Table~\ref{tab:experiment_results}, Figure~\ref{fig:strong_loss_curve}). This means the model is fitting the training set more strongly, but it is no longer improving on unseen validation text. For this reason, the best checkpoint is selected using the lowest validation loss, not the final training step.

The experiments show that MiniGPT successfully learns from the Tiny Shakespeare dataset. The baseline model is enough to verify the full training pipeline, while the stronger model gives a clear improvement in validation loss and produces better text samples. The results also show a practical limitation: increasing model capacity and training time improves performance only up to a point. After the best validation checkpoint, more training causes overfitting rather than better generalization.

\section{Qualitative Generation Analysis}

\subsection{Generated Samples}

After training, MiniGPT is evaluated qualitatively by generating text from short prompts. This is useful because validation loss shows whether the model predicts unseen text better, but generated samples show what the model has actually learned in a more readable way. Open-ended language models are often inspected using generated samples, especially when the goal is to understand the behavior of autoregressive generation \citep{radford2019language}. Decoding choices are also important, because the same trained model can produce different quality text under different sampling settings \citep{holtzman2020curious}.

The stronger MiniGPT checkpoint is used for this analysis because it gives the best validation loss. The samples are generated with \texttt{max\_new\_tokens=800}, temperature \(0.8\), and \texttt{top\_k=200}. Text is generated autoregressively: the model receives a prompt, predicts the next character, samples from the predicted distribution, appends the sampled character to the context, and repeats this process until the requested number of new characters is produced.

The following excerpts show representative generated samples. The samples are shortened for space. The full generated samples are given in the MiniGPT repo \cite{joseph2026minigpt}.

\begin{quote}
\small
\textbf{Prompt: \texttt{ROMEO:}}\\
\texttt{ROMEO:}\\
Nay, fie, I'll plead it, I have the indeed,\\
Go to the deed! I think it was a back!\\
\texttt{BRUTUS:}\\
So did let us continue them home.\\
\texttt{SICINIUS:}\\
Only, brawling not\\
The common 'twixt him, where enter'd his eyes...
\end{quote}

\begin{quote}
\small
\textbf{Prompt: \textbackslash{}n (newline character)}\\
She which she hath death seen his prey daughter:\\
It is too much not leads; but to know the right.\\
\texttt{PARIS:}\\
No, by God's deed, but I fear the sea\\
Which would have the strange for the degree.\\
\texttt{KING RICHARD III:}\\
Ay, have I been my mother's deed...
\end{quote}

The generated text is not fully meaningful, but it clearly learns some visible structure from Tiny Shakespeare dataset. It produces speaker names, line breaks, punctuation, and dialogue-like formatting. It also creates short phrases that look similar to the training dataset. This shows that the model has learned local character patterns and some formatting regularities from the dataset.

\subsection{Temperature and Top-k Sampling}

MiniGPT uses sampling rather than always selecting the most likely next character. During generation, the logits from the final time step are divided by a temperature value before applying softmax. A lower temperature makes the probability distribution sharper, so the model chooses high-probability characters more often. A higher temperature makes the distribution flatter, so the output becomes more varied but also less stable.

In this experiment, temperature has a clear effect on generation quality. With temperature $0.7$, the samples are more stable and readable. The model keeps better dialogue structure and produces fewer strange words, but the output can become more repetitive. With temperature $1.2$, the samples are more diverse and creative, but they also contain more broken spellings and unusual words such as invented names or malformed phrases. This behavior is expected because decoding strategies can strongly affect the quality and diversity of language-model output \citep{holtzman2020curious}.

MiniGPT also supports top-k sampling in the generation function. Top-k sampling keeps only the $k$ most likely next tokens before sampling. This type of sampling was also used in GPT-2 generation experiments \citep{radford2019language}. However, in this MiniGPT run, the vocabulary size is only 65 characters and the generation setting uses \texttt{top\_k=200}. Since $200$ is larger than the vocabulary size, top-k filtering does not remove any character in this specific experiment. Therefore, the main sampling factor in this experiment is temperature. The top-k option is still useful in the implementation because it makes the generation code general and easy to extend to larger vocabularies.

\begin{table}[h]
\centering
\renewcommand{\arraystretch}{1.20}
\begin{tabular}{lll}
\hline
\textbf{Setting} & \textbf{Observed behavior} & \textbf{Main trade-off} \\
\hline
Temperature = 0.7 & More stable and readable & Less diverse, more repetitive \\
Temperature = 0.8 & Balanced sample quality & Used for main samples \\
Temperature = 1.2 & More varied and creative & Less coherent, more strange words \\
Top-k = 200 & No truncation for 65-character vocabulary & Kept for generality \\
\hline
\end{tabular}
\caption{Qualitative effect of sampling settings in MiniGPT generation.}
\label{tab:sampling_effects}
\end{table}

\subsection{Observed Strengths and Failure Modes}

The main strength of the generated text is that it captures local style. MiniGPT learns to generate uppercase speaker names, colon-based dialogue markers, line breaks, punctuation, and short Shakespeare-like phrases. This is a useful result for a small character-level model, because the model is not given words directly. It has to learn spelling, word boundaries, punctuation, and formatting from characters alone.

The main failure mode is long-range meaning. The generated samples often look correct locally, but the meaning is not consistent over longer spans. Some sentences are grammatically weak, some phrases do not connect logically, and the model sometimes invents strange words. This is expected for a small character-level model trained on a narrow dataset. Character-level tokenization keeps the implementation simple, but it also makes sequences longer and makes word-level coherence harder than with subword tokenization.

The samples also show the effect of model capacity and training behavior. The stronger model generates better text than the baseline because it uses more layers, more attention heads, a larger embedding dimension, and a longer context length. At the same time, the loss curves show that training longer is not always better as discussed in Section~\ref{sec:losscurves}. The strongest validation checkpoint occurs before the final training step, and later training causes overfitting. Therefore, the best generated samples should come from the checkpoint selected by validation loss, not simply from the final model.

The qualitative analysis supports the quantitative results. MiniGPT does not generate fully coherent Shakespeare-style writing, but it learns recognizable local structure from the dataset. The model demonstrates the complete GPT-style workflow: character-level tokenization, next-token prediction, validation-based checkpoint selection, and autoregressive generation.

\section{Discussion and Limitations}

MiniGPT shows that the core GPT-style language-modeling pipeline can be rebuilt in a compact and readable PyTorch implementation. The main contribution is the complete end-to-end reconstruction of a small decoder-only language model, from raw text processing to validation-based checkpoint selection and autoregressive generation.

The first limitation is scale. MiniGPT is trained on Tiny Shakespeare dataset with a small character-level vocabulary and a relatively small model. Scaling-law studies show that language-model loss depends strongly on model size, dataset size, and training compute \citep{kaplan2020scaling}. Later work on compute-optimal training also shows that model size and training data should be scaled together under a fixed compute budget \citep{hoffmann2022training}. MiniGPT is therefore best understood as a compact implementation study, not as a competitive large-scale language model.

The second limitation is tokenization. Character-level modeling makes sequences longer and forces the model to learn words, spelling, and punctuation from individual characters. Subword tokenization methods, such as byte pair encoding, are often more efficient for larger language-model settings because they represent rare and common word pieces using a manageable vocabulary \citep{sennrich2016neural}. In the stronger configuration, the context length is 256 characters (Table~\ref{tab:training_pipeline}). Because MiniGPT uses character-level tokenization, this covers only a short passage, roughly a few dozen words, which limits long-range coherence, speaker consistency, and plot-level tracking.

The third limitation is dataset scope. Tiny Shakespeare dataset is useful because it is small, structured, and easy to train on in a notebook environment. At the same time, it is narrow in style and content. A model trained only on this dataset learns Shakespeare-like formatting, speaker names, punctuation, and local phrase patterns, but it does not learn broad language ability. The generated samples show recognizable local structure, but the meaning is often weak over longer passages.

The fourth limitation is overfitting. The stronger MiniGPT model reaches its best validation loss at step 1750 (Figure~\ref{fig:strong_loss_curve}). After that point, training loss continues to decrease, but validation loss increases. This shows that the model starts to memorize the training text more strongly instead of improving on unseen validation text. For this reason, the best model is selected used validation-based early stopping and checkpoint selection \citep{prechelt1998early}.

The fifth limitation is evaluation. This paper reports training loss, validation loss, and qualitative generated samples. These are useful for checking whether the implementation works and whether the model learns meaningful structure. However, the paper does not include large-scale benchmark evaluation, human evaluation, or comparison against pretrained models. The generated samples are therefore used only as qualitative evidence, not as a claim of strong language ability.

Finally, generation quality depends on sampling choices. Lower temperature gives more stable text, while higher temperature gives more diverse but less coherent text. This behavior is expected because decoding choices can strongly affect the quality, diversity, and repetition patterns of neural text generation \citep{holtzman2020curious}. In MiniGPT, this effect is visible even in a small character-level setting.

MiniGPT should be viewed as a reproducible implementation report. Its value is in making the GPT-style training pipeline clear and complete, from raw text to generated samples. Its limitations come mainly from the intentionally small dataset, character-level tokenization, limited model scale, and qualitative evaluation setup.

\section{Code Availability and Reproducibility}

The MiniGPT code is publicly available at \url{https://github.com/jibin10/MiniGPT}. The repository includes \texttt{MiniGPT\_Notebook.ipynb}, README instructions, a license file, and the details needed to run the experiment in Google Colab.

The implementation is written in Python using PyTorch \citep{paszke2019pytorch}. The main experiment is provided as a Jupyter notebook, so the model code, training loop, results, and generated samples can be inspected and rerun in one place \citep{kluyver2016jupyter}. The experiment uses the Tiny Shakespeare dataset with character-level tokenization \citep{karpathy2015charRNN}.

To reproduce the results, users can open \texttt{MiniGPT\_Notebook.ipynb} in Colab, select a GPU runtime, and run the notebook cells from top to bottom. The reported training times in this paper were measured on a Colab A100 GPU. The notebook sets \texttt{torch.manual\_seed(42)}. Small numerical differences may still occur because of hardware, PyTorch/CUDA versions, GPU nondeterminism, and random minibatch sampling.

\section{Conclusion}

This paper presented MiniGPT, a compact from-scratch implementation of a GPT-style autoregressive language model in PyTorch. The implementation reconstructs the end-to-end decoder-only language-modeling pipeline, from raw text processing to validation-based checkpoint selection and autoregressive generation.

On Tiny Shakespeare dataset, the baseline MiniGPT model verified that the full training pipeline works, reaching a validation loss of 1.7236 after 3000 iterations. The stronger 10.77M-parameter configuration improved the best validation loss to 1.4780 (Table~\ref{tab:experiment_results}) and generated text with recognizable Shakespeare-style local structure, including speaker names, line breaks, punctuation, and dialogue-like formatting.

MiniGPT does not propose a new Transformer architecture or a state-of-the-art language model. Its contribution is the clear and reproducible reconstruction of the GPT-style training pipeline from raw text to generated samples. The results show that even a small character-level model can learn meaningful structure when the architecture, training loop, validation process, and sampling method are implemented carefully.

\section*{Declarations}

\subsection*{Funding}
No funding was received to assist with the preparation of this manuscript.

\subsection*{Competing interests}
The author has no relevant financial or non-financial interests to disclose.

\subsection*{Data availability}
The data used in this study are publicly available through the Tiny Shakespeare dataset used in Karpathy's character-level language modeling examples \citep{karpathy2015charRNN}.

\subsection*{Code availability}
The code is available at \url{https://github.com/jibin10/MiniGPT}.

\bibliographystyle{unsrtnat}
\bibliography{references}  






\end{document}